\documentclass[10pt,twocolumn,letterpaper]{article}

\usepackage{wacv}
\usepackage{times}
\usepackage{epsfig}
\usepackage{graphicx}
\usepackage{amsmath}
\usepackage{amssymb}
\usepackage[labelformat=empty]{subfig}
\usepackage{caption}
\usepackage{booktabs}
\usepackage{makecell}
\usepackage{comment}
\usepackage{afterpage}
\usepackage{color}

\def\wacvPaperID{229} 

\wacvfinalcopy 

\ifwacvfinal
\def\assignedStartPage{1} 
\fi

\ifwacvfinal
\usepackage[breaklinks=true,bookmarks=false]{hyperref}
\else
\usepackage[pagebackref=true,breaklinks=true,colorlinks,bookmarks=false]{hyperref}
\fi

\ifwacvfinal
\else
\fi

\begin{document}

\title{Driving among Flatmobiles: Bird-Eye-View occupancy grids from a monocular camera for holistic trajectory planning }

\author{
   Abdelhak Loukkal\thanks{Renault S.A.S, 1 av. du Golf, 78288 Guyancourt, France.} \ \thanks{Sorbonne universités, Université de technologie de Compiègne, CNRS, Heudiasyc, UMR 7253, Compiègne, France.}
   \and
   Yves Grandvalet\footnotemark[2]
   \and
   Tom Drummond\thanks{Department of Electrical and Computer Systems Engineering, Monash University, Clayton, VIC, Australia}
   \and
   You Li\footnotemark[1]
}
\maketitle

\begin{abstract}

Camera-based end-to-end driving neural networks bring the promise of a low-cost system that maps camera images to driving control commands. These networks are appealing because they replace laborious hand engineered building blocks but their black-box nature makes them difficult to delve in case of failure. Recent works have shown the importance of using an explicit intermediate representation that has the benefits of increasing both the interpretability and the accuracy of networks' decisions. Nonetheless, these camera-based networks reason in camera view where scale is not homogeneous and hence not directly suitable for motion forecasting. In this paper, we introduce a novel monocular camera-only holistic end-to-end trajectory planning network with a Bird-Eye-View (BEV) intermediate representation that comes in the form of binary Occupancy Grid Maps (OGMs). To ease the prediction of OGMs in BEV from camera images, we introduce a novel scheme  where the OGMs are first predicted as semantic masks in camera view and then warped in BEV using the homography between the two planes. 
The key element allowing this transformation to be applied to 3D objects such as vehicles, consists in predicting solely their footprint in camera-view, hence respecting the flat world hypothesis implied by the homography.

\end{abstract}
\vspace{-1.125ex}

\section{Introduction}

\begin{figure}[t!]
\includegraphics[width=0.48\textwidth]{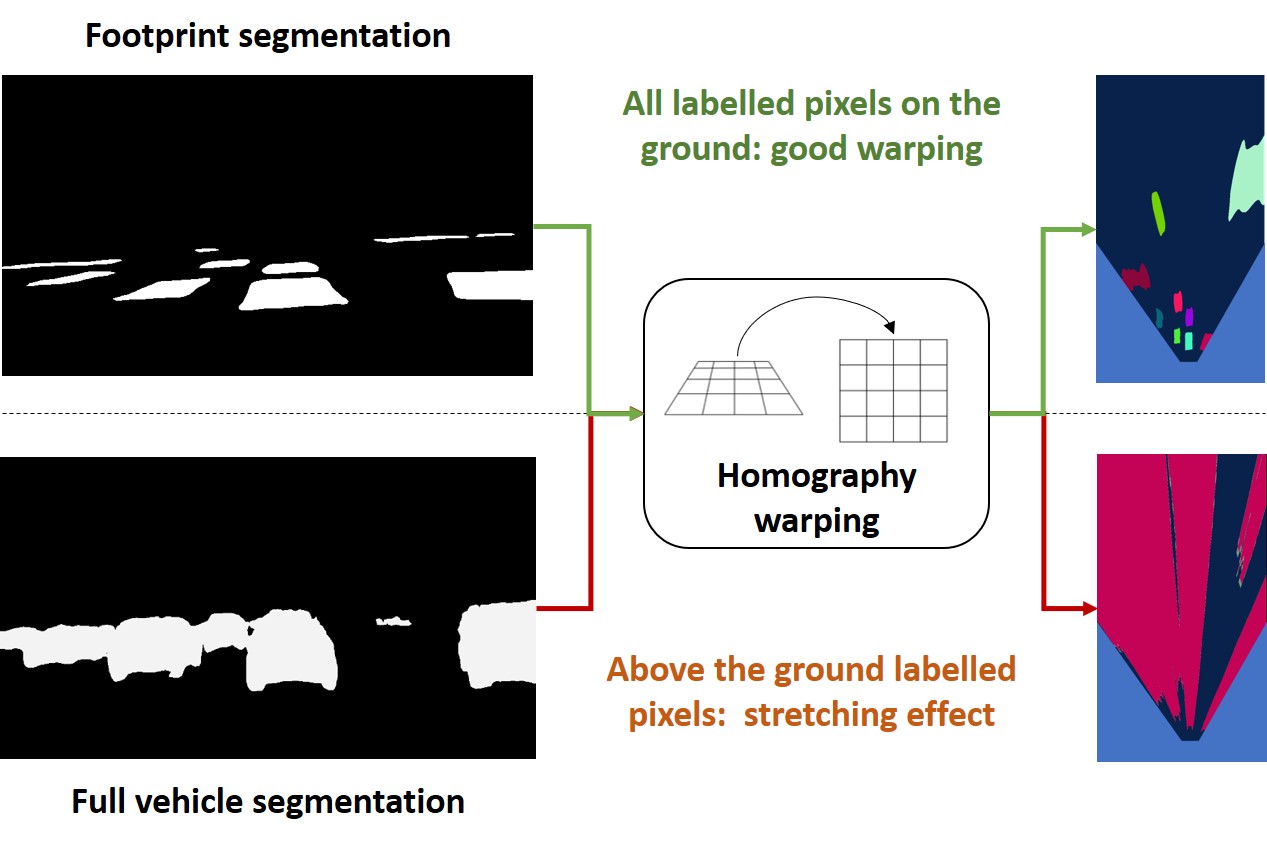}
\caption{Applying a planar homography to a full segmentation mask causes the pixels located above the ground plane to be stretched in BEV. Our footprint segmentation respects the  flat  world hypothesis implied by the homography hence avoiding the deformation caused by above the ground pixels.} \label{is}
\end{figure}

Autonomous driving systems are usually composed of carefully engineered modular building blocks. Each building block being developed separately, this makes the whole system more interpretable but can also lead to an accumulation of errors along the pipeline. End-to-end driving CNNs \cite{Pomerleau1988ALVINNAA,DBLP:journals/corr/BojarskiTDFFGJM16,Codevilla2017EndtoEndDV,DBLP:journals/corr/abs-1906-03199} consist in neural networks that take raw sensor data as input, or other modalities such as the depth or the optical flow, and output  control commands, for example steering wheel angles. This kind of approach is promising but lacks interpretability because it skips all the building blocks of the autonomous driving pipeline \cite{Zeng_2019_CVPR}.
Mid-to-mid driving \cite{DBLP:conf/rss/BansalKO19} is a more recent approach to this problem that instead of taking images as input takes intermediate representations of the driving scene provided by the perception building block. In the middle ground of these two approaches lie mediated perception approaches \cite{DBLP:conf/iccv/ChenSKX15,Sauer2018ConditionalAL,DBLP:conf/iccv/KimC17,DBLP:conf/corl/MuellerDGK18,Li2018RethinkingSM} that take raw sensor data as input, have an explicit intermediate representation of the observed scene and then forecast motion based on this representation. However, none of the existing monocular camera based end-to-end approaches leverage a Bird-Eye-View (BEV) intermediate representation when it seems to be the best candidate for motion forecasting as the scale in this plane is homogeneous and the size of the observed objects is invariant to their distance to the sensor. 

In this paper, we introduce a novel holistic end-to-end trajectory planning network that takes a sequence of monocular images as input, leverages a BEV mediated perception strategy and plans the trajectory of the ego vehicle knowing its destination position at a 3s horizon. The intermediate representation of our network comes in the form of two binary Occupancy Grid Maps (OGMs) in BEV, one giving the drivable area and the other one the occupancy of the vehicles in the scene.
As it is not straightforward to directly obtain BEV outputs from a camera plane input while preserving the receptive field, the first stage of our network first outputs these OGMs in camera view then warps them in BEV using the planar homography between the camera plane and the BEV plane. However, a planar homography relates a transformation between two planes, so we introduce the Flatmobile representation to warp the vehicles from camera view to BEV without the ``stretching" effect observed in Fig.~\ref{is}. This simple yet very effective representation consists in segmenting the footprint of the vehicles, i.e., the ``ground" face of their 3D bounding boxes in camera view, so that the flat world assumption is respected when warping these masks in BEV. In an end-to-end fashion, an encoder-decoder LSTM is fed with these OGMs, the past trajectory and the destination position of the ego vehicle to produce the future trajectory. 

In summary, the contributions of this work are two-fold:
\begin{itemize}
    \item The main contribution is to propose a novel learning protocol to learn BEV Occupancy Grids from a monocular camera;
    \item The relevance of this protocol is shown by its integration into a holistic end-to-end trajectory planning pipeline that takes advantage of these occupancy grid maps as an intermediate representation.
\end{itemize}

The gain in performance on both tasks is demonstrated on the nuScenes dataset \cite{nuscenes2019}.
\begin{figure*}[t]
\begin{center}
\includegraphics[width=1\textwidth]{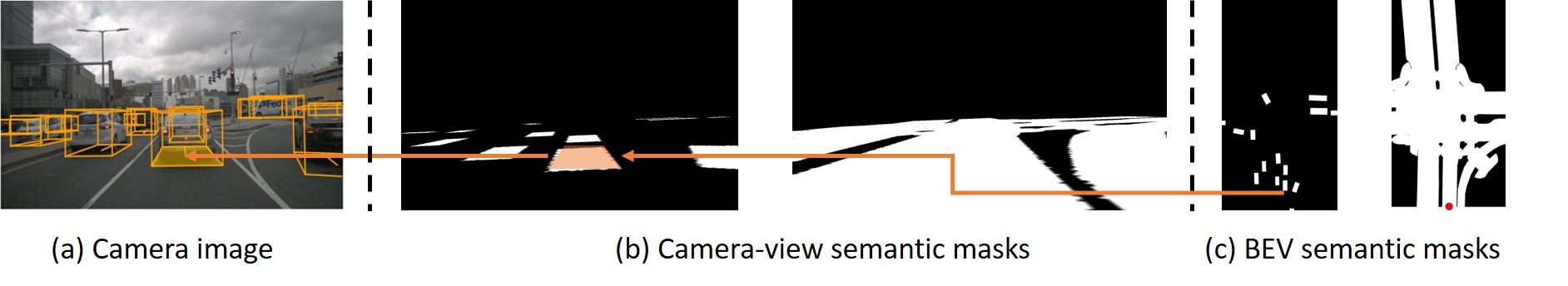}
\end{center}
\caption{For each camera frame $(a)$, a BEV map of size $100\times55$ (in meters) is extracted as a binary semantic mask. A vehicle binary mask with the same scale and size is generated from the ground-truth 3D bounding boxes $(c)$, where the red dot indicates the ego vehicle. 
These two masks are projected in camera view $(b)$ using the homography between the two planes. Better viewed in color.} \label{data}
\end{figure*}
\section{Related work}
\paragraph{Occupancy grid maps:}
Occupancy grid maps \cite{DBLP:journals/corr/abs-1304-1098}
represent the spatial environment of a robot and reflect its occupancy as a fine-grained metric grid. These grid maps have been widely used to model the environment of indoor and outdoor mobile robots as well as automotive systems. Once acquired, they can be used for various tasks such as path planning. OGMs can be acquired with range sensors like LiDAR or RADAR, but also from RGB-D cameras \cite{article}, stereo cameras \cite{DBLP:journals/sensors/LiR14}, or from the fusion of multiple sensors \cite{fusi}. For example, semantic OGMs are predicted from a LiDAR and a monocular camera thanks to deep learning and Bayesian filtering in \cite{DBLP:conf/iros/Erkent0LGR18}.  Taking only monocular camera images as input, a Variational Encoder-Decoder (VED) can predict semantic OGMs \cite{Lu2018MonocularSO}. However, this approach leverages poor ground-truth OGMs obtained by warping semantic segmentation masks using the associated disparity maps.
The Orthographic Feature Transform (OFT) network \cite{roddick2018orthographic}, whose original purpose is to output 3D bounding boxes, consists in projecting the features extracted in camera view to an orthographic space. To do so, voxel-based features are generated by accumulating camera-view features; then these features are collapsed along the vertical dimension to obtain features in an orthographic plane. In addition to bounding boxes coordinates and dimensions, this network also outputs a confidence map in BEV which can be assimilated to an occupancy grid map. To the best of our knowledge, OFT and VED are the only comparable approaches to take monocular camera images as input and output a dense OGM-like grid. 
\vspace{-1.125ex}
\paragraph{Behavior cloning based ego-motion forecasting:}
Imitation learning consists in reproducing a desired behavior based on expert demonstrations. Behavior cloning consists in learning a policy that reproduces the desired behavior by learning a direct mapping from states to actions. Autonomous Land Vehicle In a Neural Network (ALVINN) \cite{Pomerleau1988ALVINNAA} was a pioneering work on ego-motion forecasting that used behavior cloning and a neural network to accomplish lane following. More recently, \cite{DBLP:journals/corr/BojarskiTDFFGJM16} introduced an end-to-end driving network that maps camera images to steering commands. A similar approach was adopted in \cite{Codevilla2017EndtoEndDV} with the difference that the steering commands are conditioned with high-level commands, e.g., turn left. Also adopting conditional imitation learning, \cite{DBLP:journals/corr/abs-1906-03199} uses multi-modal inputs and explores different fusion schemes. In our holistic end-to-end network, we condition our output with the destination position which makes it a trajectory planning set-up.

Another body of work incorporates an intermediate representation in the form of affordances \cite{DBLP:conf/iccv/ChenSKX15,Sauer2018ConditionalAL}, attention maps \cite{DBLP:conf/iccv/KimC17} or semantic segmentation masks \cite{DBLP:conf/corl/MuellerDGK18,Li2018RethinkingSM}. These approaches address the main flaw of the end-to-end approach by making the results interpretable by a human examinator. Instead of learning an intermediate representation and learning motion from this representation, \cite{Xu2016EndtoEndLO} learned to forecast motion from a sequence of input images and learned the semantic segmentation of these input images as a side task invoking privileged learning.  However, all these camera-based solutions reason in camera view whereas it seems to be more suitable to forecast motion in BEV where the size of objects does not depend on their position in the image. The network developed in this paper takes only monocular images as input but has a BEV intermediate representation that is used for motion forecasting. 

ChauffeurNet \cite{DBLP:conf/rss/BansalKO19} adopts a mid-to-mid driving model that uses the BEV output of a perception module as input to predict a trajectory. Using a mid-level representation as input allows to augment the training  data with synthetic worst case scenarios hence improving the performance of the network in a real-world scenario. Mid-to-mid driving is also adopted by the privileged learner of \cite{chen2019lbc}. Alleviating the burden of manually designed planning cost functions\cite{Paden2016ASO}, \cite{Zeng_2019_CVPR} introduces an end-to-end motion planner that takes a 3D point cloud and an HD map as input and has an intermediate interpretable representation in the form of 3D detections and their predicted trajectories. 

\section{End-to-end planning with BEV mediated perception}

Since our method requires certain annotations in the nuScenes dataset \cite{nuscenes2019}, we start by describing them. They will be used directly or indirectly as targets in the learning process of our holistic end-to-end network. 

\subsection{Data}
Experiments are conducted on the nuScenes dataset \cite{nuscenes2019}, which records the measurements of a complete suite of sensors: 6 cameras, 32-channels LIDAR, long-range radars. The whole dataset comprises 40\,000 annotated frames but for intellectual property reasons we are currently limited to the preview version, with 3\,340 annotated frames from five driving sequences in Boston and 81 sequences in Singapore. Each driving sequence lasts around 20s and images are acquired at a framerate of 2Hz. The two most important features of nuScenes  for this paper are  the availability (i) of 3D bounding boxes, (ii) of layers of the drivable area are also provided as binary semantic masks, where each pixel corresponds to $0.1\times0.1$ square meters. 

\vspace{-1.125ex}
\paragraph{Ground-truth OGMs:}
For each of the considered frames, a map portion of size $1\,000\times550$ (that is, 100m$\times$55m) is extracted with the ego vehicle positioned at $(1000,300)$ from the origin at the top left corner of the map portion. This map portion is rotated according to the ego vehicle heading angle such that it always faces forward. 
Given the 3D position of each vehicle in the scene, it is possible to draw its ground-truth bounding boxes on a blank canvas aligned with the map portion, with the correct size and position. These two binary semantic masks, depicted in the right-hand side of Fig.~\ref{data}, can also be viewed as OGMs with grid cells of size 0.1m $\times$ 0.1m.

\vspace{-1.125ex}
\paragraph{Ground-truth trajectories:}
In addition to the semantic masks, the past and future trajectories are also pre-processed. The past and future positions are measured relatively to the current position of the ego vehicle. The past trajectory is defined as the 6 previous positions, with a 0.5s time-step. The future trajectory is defined in the same way, with the sixth point defined as the destination and the 5 intermediate points as ground truth for  motion planning. After removing the frames that do not have 6 previous and future positions, the dataset is split in 1917 training samples and 415 testing samples from held-out sequences.

\vspace{-1.125ex}
\paragraph{Homography estimation:}

A key input of our holistic end-to-end network is the homography matrix between the camera plane and the OGM plane. The homography between the two planes maps a point in the camera plane $(u,v,1)^{T} $ to a point in the BEV plane $(x,y,1)^{T}$ such that:
\begin{equation}
\centering 
\begin{bmatrix}x\\ y\\ 1
\end{bmatrix} = 
\begin{bmatrix}
h_{11} & h_{12}  & h_{13}\\ 
 h_{21}&h_{22}  & h_{23}\\ 
 h_{31}& h_{32} & h_{33}
\end{bmatrix}
\begin{bmatrix}u\\ v\\ 1\end{bmatrix}
\enspace.
\end{equation}
Our framework relies on the availability of the homography matrix at each frame. For doing so, we build a training set with ground-truth homographies to predict homographies from camera frames.

\begin{figure}[t]
\begin{center}
\includegraphics[width=8.3cm]{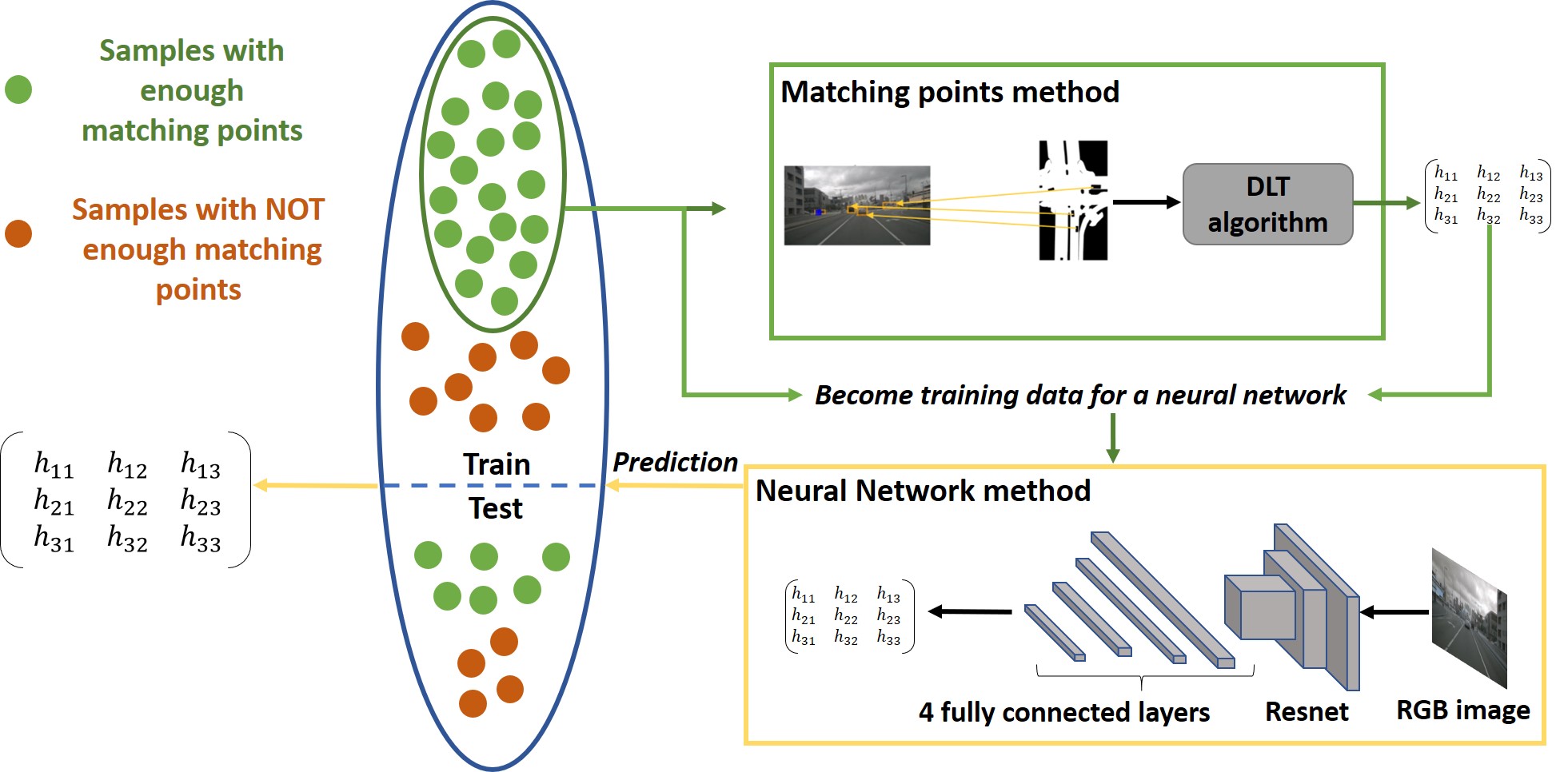}
\caption{For training images with enough matching points, the homography is computed with the matching method. These samples become training data for a neural network that learns to estimate the homography from RGB input images. This trained network is then used to pre-compute the homographies of all the samples in the dataset. Better viewed in color.}
\label{nu}
\end{center}
\end{figure}

We take advantage of the annotated 3D bounding boxes to get corresponding points in the BEV and the camera planes. The pixel positions of the 3D bounding boxes ``ground" face corners in the camera image are matched with their 3D position in the BEV plane (see Fig.~\ref{data}), and the homography matrix is obtained using the \texttt{getPerspectiveTransform()} function of OpenCV that applies the Direct Linear Transformation (DLT) algorithm \cite{Hartley:2003:MVG:861369}. 
Not all training samples contain enough matching points, so only a subset of the original training set is available for fitting a network predicting homographies  (see Fig.~\ref{nu}).
The homography network is composed of a ResNet-18 encoder and 4 fully connected layers with the fourth layer outputting 9 values corresponding to the elements of the homography matrix. The homography matrix contains 4 rotational terms and 2 translation terms. The difference in magnitude between these terms must be taken into account during training. The authors of \cite{DBLP:journals/corr/DeToneMR16} circumvent this issue by predicting 4 matching points instead of the homography matrix elements as there is a one-to-one correspondence between the two representations; we simply rescale the rotational and translation terms with a constant factor such that all the elements of the homography have the same magnitude. The L2 loss function is then used as the training criterion.

For each frame in the dataset, the predicted homography is used to warp the drivable area and vehicle OGMs in the camera plane (cf. Fig.~\ref{data}). These warped masks are used as ground truths in the network described in Fig.~\ref{homo}.
Each sample has a sequence of 6 input camera images separated by 0.5s, the camera view semantic masks for every image in the sequence, the homography matrices for every mask in the sequence, the past and future trajectories.

\subsection{Footprint segmentation: OGMs from a monocular camera}
\paragraph{Model formulation:}
\begin{figure*}[t]
\begin{center}
\includegraphics[width=0.95\textwidth]{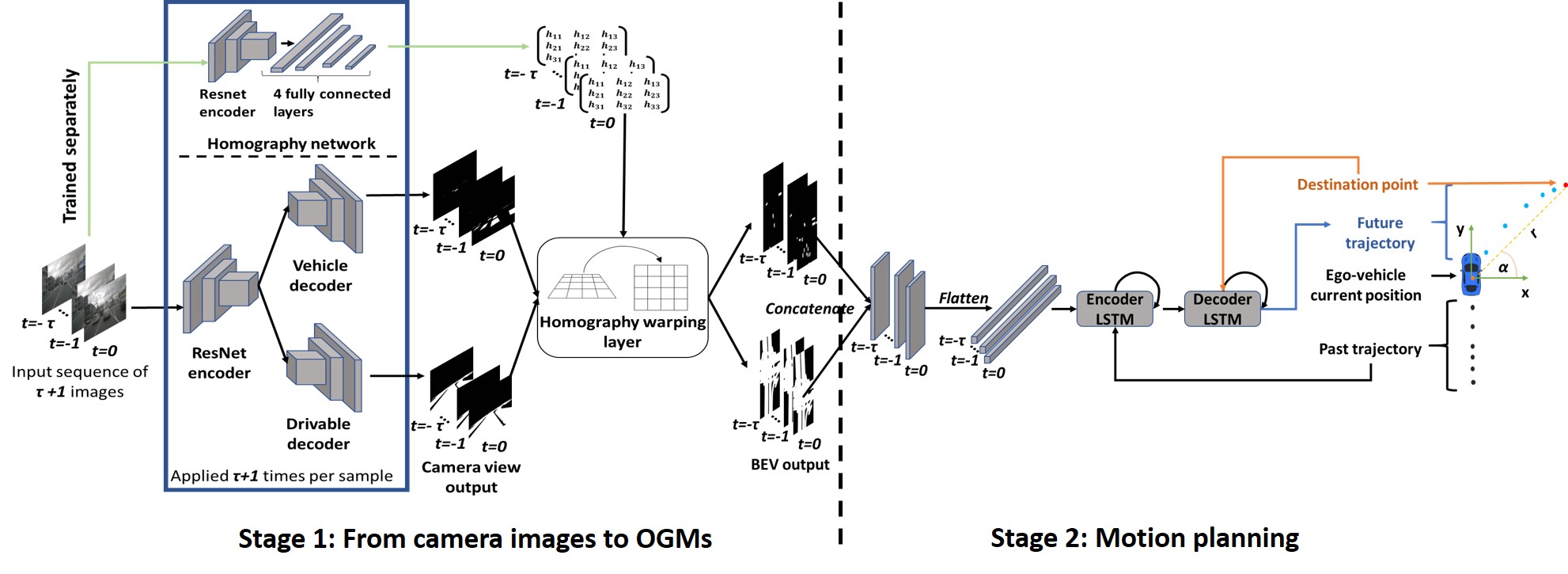}
\caption{Two-stage network for end-to-end trajectory planning from a monocular camera with intermediate BEV OGM outputs. Better viewed in color.}
\label{homo}
\end{center}
\end{figure*}
The overall approach consists of two stages, as depicted in Fig.~\ref{homo}. The first stage generates OGMs from the input sequence of monocular images. These OGMs' cells can take two states, occupied or free.
Two OGMs are considered: one that gives an information about the drivable area, and the other one about the position of the vehicles in the observed scene. The considered OGMs being homologous to binary semantic masks, their estimation is formulated as the semantic segmentation of the drivable area and the vehicles in the scene. State of the art semantic segmentation CNNs come in an encoder-decoder configuration where the input and the output of the network have the same aspect ratio and are in the same geometric plane. The ultimate goal of this stage is to output BEV semantic masks for each image of the input sequence. However, it is not straightforward to directly output BEV masks from a camera plane input as the receptive field of a pixel in the BEV output would not match the region of the image that is responsible for its processing. 
Hence, the semantic masks are first outputted in the camera plane. ResNet-101 \cite{DBLP:conf/cvpr/HeZRS16} is used as the encoder and  Deeplab v3+ \cite{DBLP:conf/eccv/ChenZPSA18} as the decoder. 
Deeplab v3+ combines A trou Spatial Pyramid Pooling (ASPP) \cite{DBLP:conf/eccv/HeZR014} and a convolutional decoder module to benefit from rich contextual information and better object boundaries delineation. 
Two decoding heads output the drivable area and vehicle semantic masks in camera view. The encoder-decoder network is applied $t$ times where $t$ is the number of images in the input sequence. 
The final step of this stage of the network is to warp the output masks in BEV using a perspective warping layer \cite{Arraiy2018} with the homography matrix between the camera plane and the BEV plane.
However, vehicles are 3D objects and projecting regular semantic masks in BEV leads to the 
``stretching" effect observed in Fig.~\ref{is}. To cope with this deformation, we introduce the Flatmobile representation to warp the vehicles from camera view to BEV. This simple yet very effective representation consists in segmenting the footprint of the vehicles, i.e., the ``ground" face of their 3D bounding boxes in camera view. By doing so, only pixels located on the ground surface (assumed to be locally flat) are segmented and warped in BEV which respects the planar world hypothesis required by the homography.
The homography for each sample is obtained with a homography estimation network trained separately from the end-to-end network but on samples from the same training set, see Fig.~\ref{nu}. The output size of the decoding heads is $s$ times smaller than the original camera image so the homography needs to be re-scaled as following:

\begin{equation}
    H_s = S\cdot H \cdot S^{-1}
    \enspace,   \  \text{with} \enspace
    S = 
\begin{bmatrix}
 s & 0 & 0 \\ 
 0 & s & 0 \\ 
 0 & 1 & 0
\end{bmatrix}
\enspace,
\end{equation}
where $H_s$ is the re-scaled homography, $H$ the original homography and $s$ is the (fixed) scaling factor accounting for the ratio of image to decoder output size.

\paragraph{Cost function and learning:}
We initially tried to make the most of ground-truth signals by computing two losses, one in the camera plane and the other one in the bird-eye view plane.  However, we eventually train from the camera plane only, since the bird-eye view loss diverged. This is due to the fact that errors in the camera plane are amplified in the BEV plane.
We chose to predict separately a binary OGM for the vehicles and another one for the drivable area, since our problem is best formalized as a multi-label classification  problem rather than a multiclass classification problem. 
Indeed, each cell can belong to one of the four classes: 
\{vehicle, drivable, vehicle and drivable, none\}.
The overall loss of this stage of the network, denoted $\mathcal{L}_{perception}$, is defined as the sum of two binary cross-entropy losses: one for the drivable area, and the other one for the vehicle occupancy.

\begin{table*}
\caption{OGMs evaluation by $IoU$ (in \%). Full refers to the whole OGMs, close focuses on a region of 50m ahead of the ego vehicle and 10m on each side, far on the region farther than 50m ahead of the ego vehicle.}
\label{iou}
\centering
\begin{tabular}{@{}l*9{c}@{}}
	\toprule
     & \multicolumn{4}{c}{IoU drivable} & \hspace*{1.5ex} & \multicolumn{4}{c}{IoU vehicles} \\ 
	 \cline{2-5} \cline{7-10} \\[-2.5ex]
            & Camera & \multicolumn{3}{c}{Bird-Eye}  && Camera & \multicolumn{3}{c}{Bird-Eye} \\
	  \cline{3-5} \cline{8-10} \\[-2.5ex]
      Range & Full   & Full   & Close   & Far   && Full  & Full   & Close   & Far   \\[-.5ex] \midrule
VED \cite{Lu2018MonocularSO} & NA & 60.1 & 48.6 & 70.3  && NA     & 48.9 & 47.1 & 50.0 \\
OFT \cite{roddick2018orthographic} & NA & 58.6 & 41.2 & \textbf{71.4}  && NA     & 60.0 & 54.6 &  \textbf{55.9} \\
Ours   & 94.5   & \textbf{61.1}   & \textbf{92.4}    & 40.5  && 79.3   & \textbf{65.4} & \textbf{80.0} & 49.7 \\
\midrule
BEV LiDAR & NA & 86.5  & 68.5  & 90.9  && NA  & 69.7 & 69.9 & 49.8 \\
 \bottomrule
\end{tabular}
\end{table*}
\subsection{Encoder-decoder LSTM for trajectory planning}
\paragraph{Model formulation:}
\begin{figure}[t!]
\begin{center}
\includegraphics[width=8.3cm]{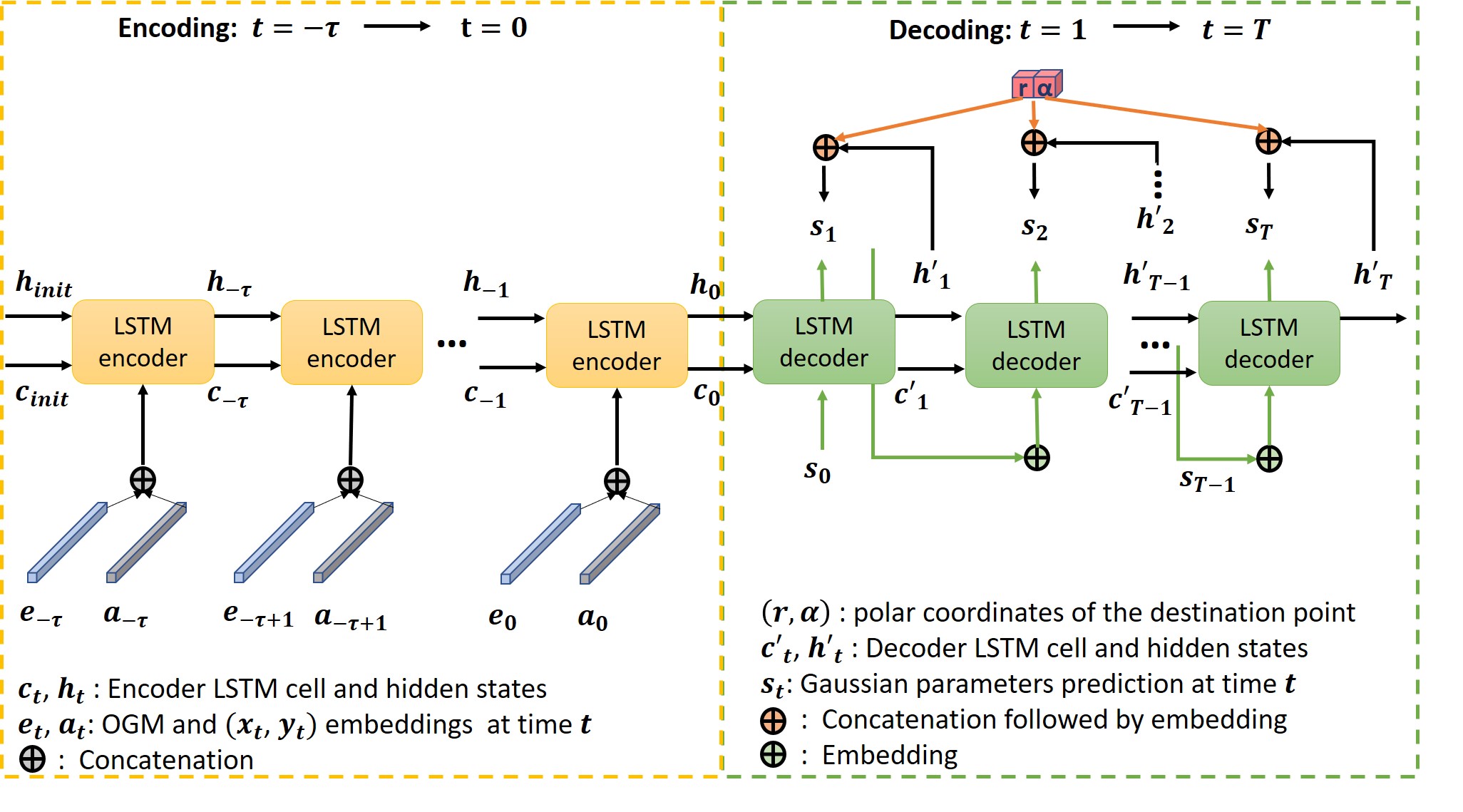}
\caption{Encoder-decoder LSTM for trajectory planning. Better viewed in color.}
\label{lstm}
\end{center}
\end{figure}
The second stage of the network implements mid-to-end trajectory planning. The two sequences of drivable area and vehicle BEV masks are concatenated along the channel dimension at each time step and flattened to obtain a sequence of feature vectors 
$f_{-\tau}, ..., f_0$. 
These feature vectors become input to an encoder-decoder LSTM \cite{cho-etal-2014-learning}. The encoder embeds the sequence of feature vectors of length 
$\tau+1$ 
and processes it introducing the following recurrence:
\begin{equation}
\label{enco}
\begin{split}
& e_t = \phi_e (f_t; W_e)\\
& a_t = \phi_a ((x_t, y_t); W_a) \\
& h_t = LSTM (h_{t-1}, concat(e_t,a_t); W_l)
\enspace,
\end{split}
\end{equation}
where $(x_t, y_t)$ is the position of the ego vehicle at time $t$, $\phi_e$ and $\phi_a$ are the embedding functions, $W_e$ and $W_a$ are the embedding weights, $h_t$ and $W_l$ are respectively the hidden state and the weights of the encoder LSTM. 

The final cell state $c_0$ and hidden state $h_{0}$ of the encoder that summarizes the input OGM sequence are fed to the decoder as its initial cell and hidden states. The decoder then recursively outputs the sequence of future positions: 

\begin{equation}
\label{deco}
\begin{split}
& a^{\prime}_t = \phi^{\prime}_a (s_{t-1};W^{\prime}_a) \\
& h^{\prime}_t = LSTM (h^{\prime}_{t-1},a^{\prime}_t; W^{\prime}_l)\\
& s_{t} = \phi_s (concat(h^{\prime}_t,(r,\alpha)); W_s) 
\enspace,
\end{split}
\end{equation}
where $\phi^{\prime}_a$ and $\phi_s$ are the embedding functions, $W^{\prime}_a$ and $W_s$ are the embedding weights, $(c^{\prime}_t$, $h^{\prime}_t)$ and $W^{\prime}_l$ are respectively the cell/hidden states and the weights of the decoder LSTM, $s_t$ the estimated future position at time $t$ and $(r, \alpha)$ the polar coordinates of the destination point. 
The prediction of the future positions $s_{1:T}$ is expressed in Cartesian coordinates, whereas the destination point is expressed in polar coordinates, for encouraging solutions that depart from a simple direct interpolation, thereby enhancing the impact of different design choices. 
\paragraph{Cost function and learning:}
At each time-step during decoding (i.e., planning) time, the LSTM predicts the 
distribution of the
future position of the ego vehicle as developed in \eqref{deco}. Similar to \cite{Alahi_2016_CVPR}, the output $s_t$ of the encoder-decoder LSTM module predicts the parameters of a bi-variate Gaussian distribution characterized by its mean $\mu_t = (\mu^x_t,\mu^y_t)$ and its covariance matrix parameterized by the standard deviations $\sigma_t= (\sigma^x_t,\sigma^y_t)$ and the correlation $\rho_t$.

The predicted position of the ego vehicle at time $t$ is given by $
(x_t,y_t) \sim \mathcal{N}(\mu_t,\sigma_t, \rho_t)$.
The parameters of the encoder-decoder LSTM module are learned by minimizing the negative log-likelihood of the Gaussian distribution: 

\begin{equation}
\mathcal{L}_{motion} = -\sum_{t = 1}^{T} \log (\mathbb{P}(x_t,y_t|\mu_t,\sigma_t, \rho_t)) 
\enspace.
\end{equation}
For the holistic end-to-end network, the final loss function is a linear combination of the perception loss and the motion loss:

\begin{equation}
 \mathcal{L}_{total}=\alpha \,\mathcal{L}_{perception}+\mathcal{L}_{motion}  \enspace,
\end{equation}
where $\alpha$ is empirically set to $0.1$.

\section{Experimental evaluation}
\subsection{Evaluation metrics}
The Average Displacement Error ($ADE$) corresponds to the average Euclidean distance between the predicted trajectory and the ground-truth one: $ADE = \frac{1}{N\,T}\sum_{i = 1}^{N}\sum_{t=1}^{T}\big\| \widehat{Z}_{i}^{t} - Z_{i}^{t} \big\|_2$,
where $N$ is the number of samples, $T$ is the number of prediction timesteps,  $Z_{i}^{t}$ are the $i$th ground-truth coordinates at time step $t$
and $\widehat{Z}_{i}^{t}$ are their predictions.
As the OGM prediction amounts to semantic segmentation, the classical Intersection over Union ($IoU$) metric is also adopted.
\subsection{Occupancy grid maps}
\paragraph{Evaluation baselines}
The performance of the first stage, OGMs from a monocular camera, is evaluated against the following baselines:
\begin{itemize}
    \item VED \cite{Lu2018MonocularSO}: Multi-task training led to poor results so this model was trained to output the vehicles and drivable area OGMs separately. 
    \item OFT \cite{roddick2018orthographic}: this model was trained with a binary cross-entropy loss to solely output the occupancy grid maps.
    \item BEV-LiDAR: The available LiDAR point clouds were processed to obtain BEV 3-channels images that encode the distance to the LiDAR, the height and the intensity. A Deeplab v3+ was then trained to take these images as input and output the vehicles and drivable area OGMs.
\end{itemize}
\paragraph{Training setup}
In this evaluation, the models are compared on the whole preview dataset. 
The dataset is split into 77 training sequences (2982 frames) and 9 held-out testing sequences (358 frames). 
All networks are trained over 100 epochs with Stochastic Gradient Descent (SGD) and a batch size of 8, with learning rate of $10^{-7}$ for OFT and $10^{-3}$ for our network and VED. 
The grid in OFT is set to $100\times55$ (in meters) with resolution of 1m per pixel.
\paragraph{Quantitative results}
\begin{figure*}[t!]
\begin{center}
\begin{tabular}{@{}p{0.38\textwidth}*4{p{0.12\textwidth}}@{}}

{\includegraphics[width=0.38\textwidth]{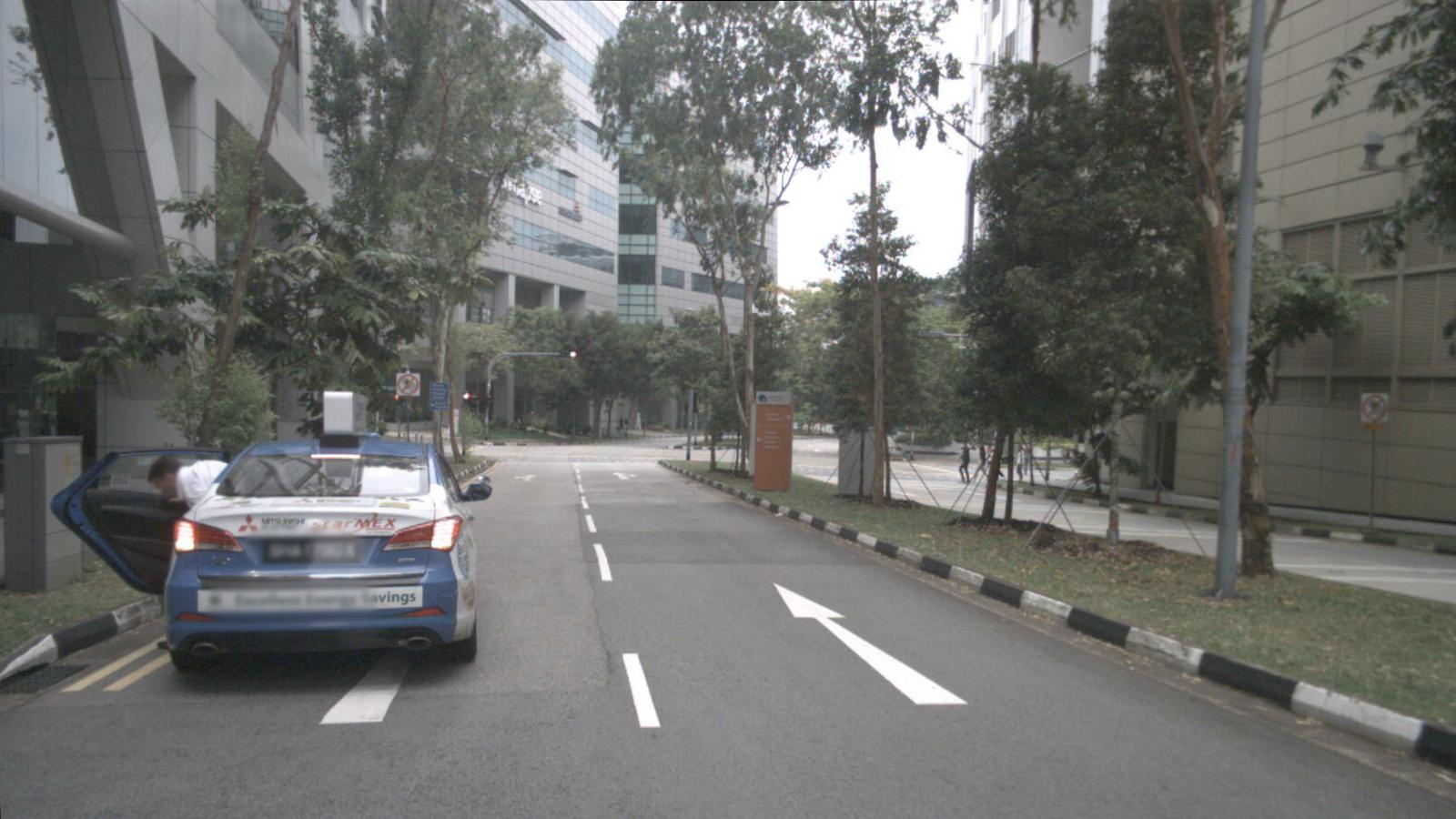}} &
{\includegraphics[width=0.12\textwidth]{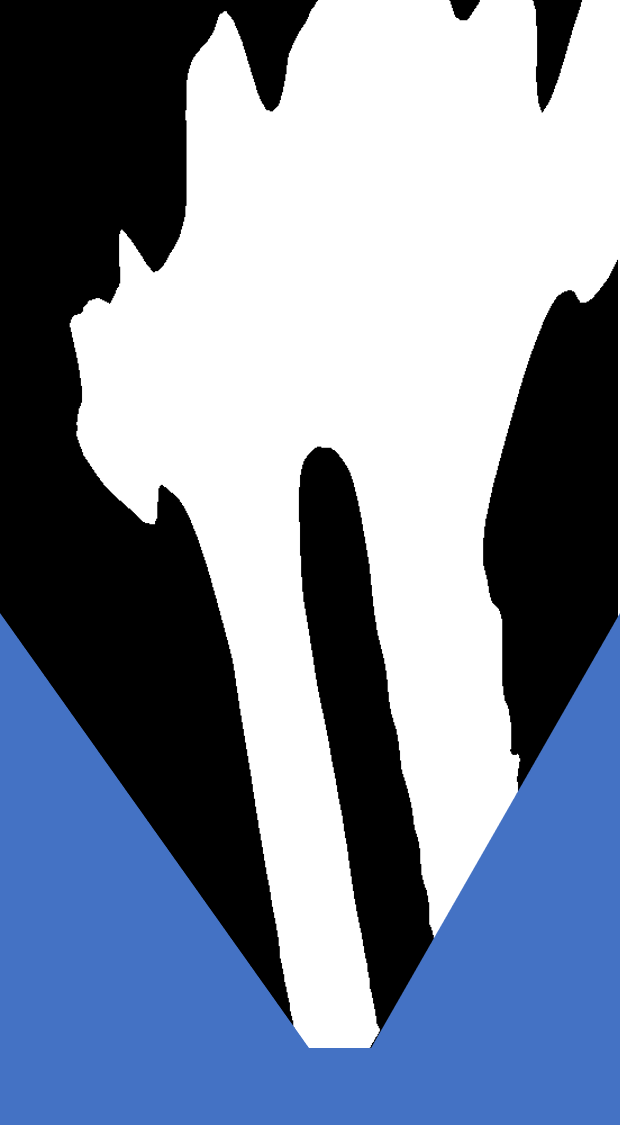}}&
{\includegraphics[width=0.12\textwidth]{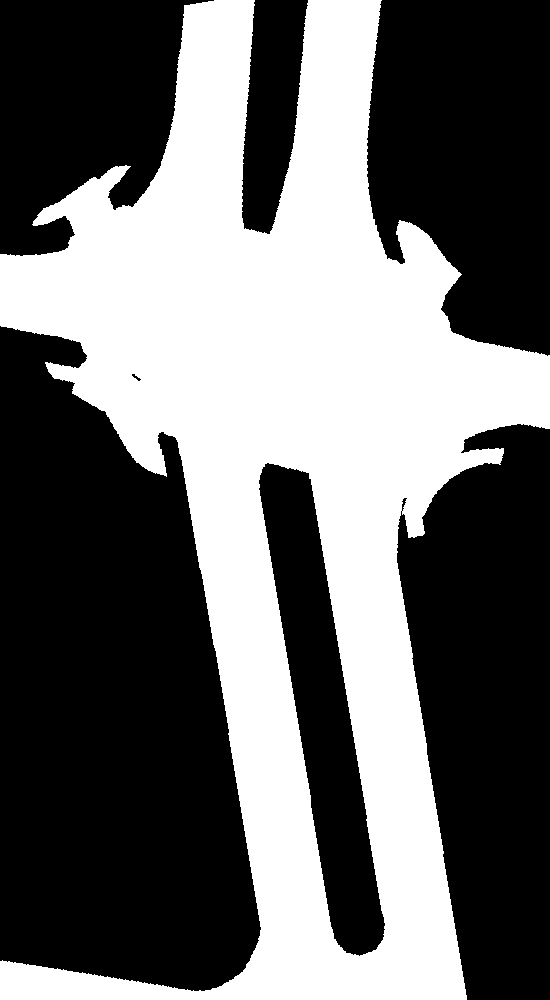}} &
{\includegraphics[width=0.12\textwidth]{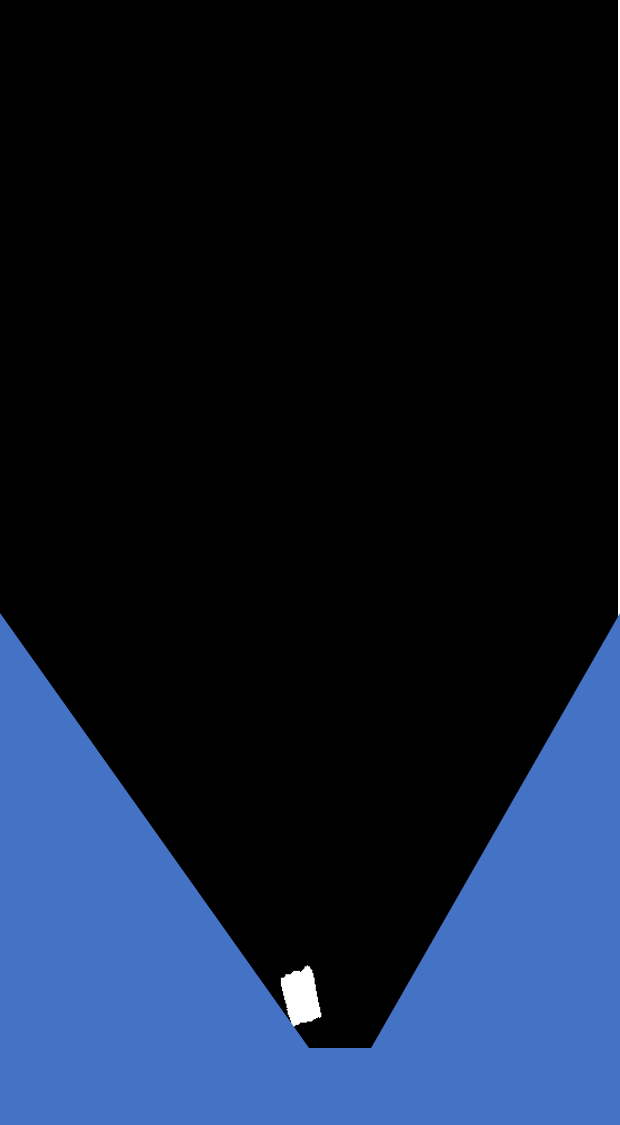}} &
{\includegraphics[width=0.12\textwidth]{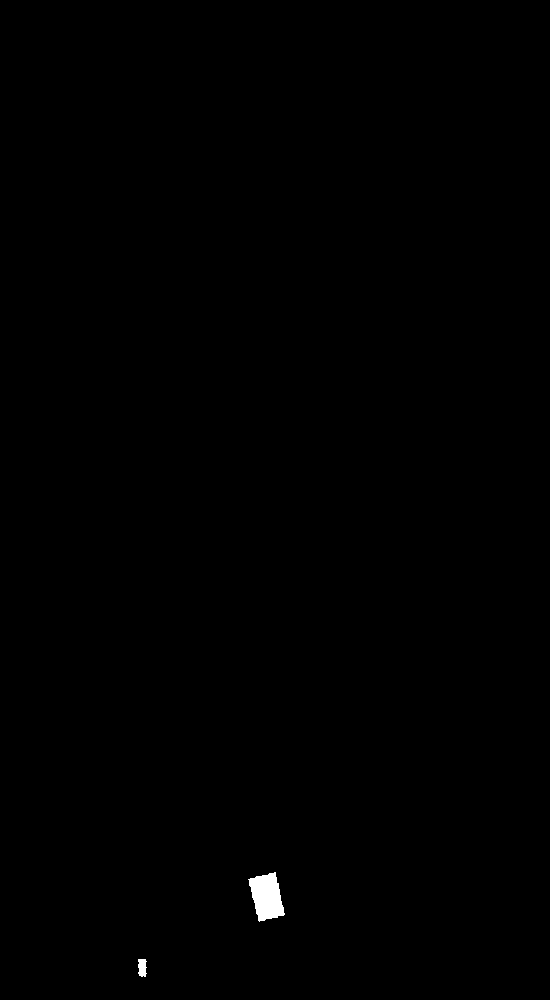}} \\
{\includegraphics[width=0.38\textwidth]{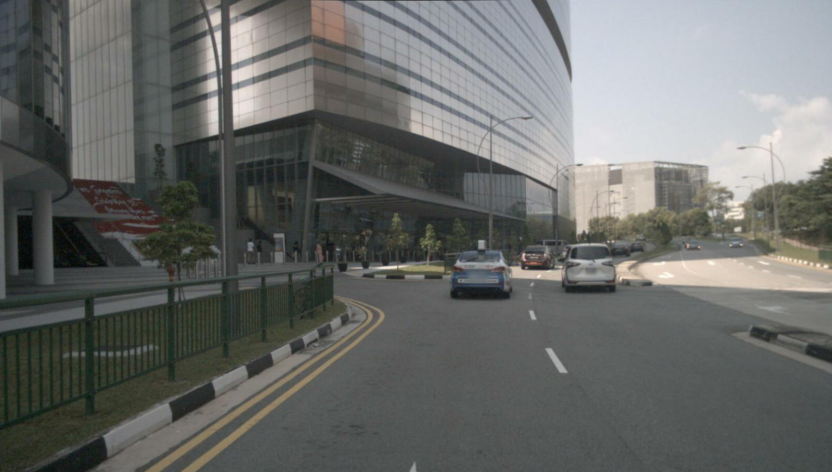}} &
{\includegraphics[width=0.12\textwidth]{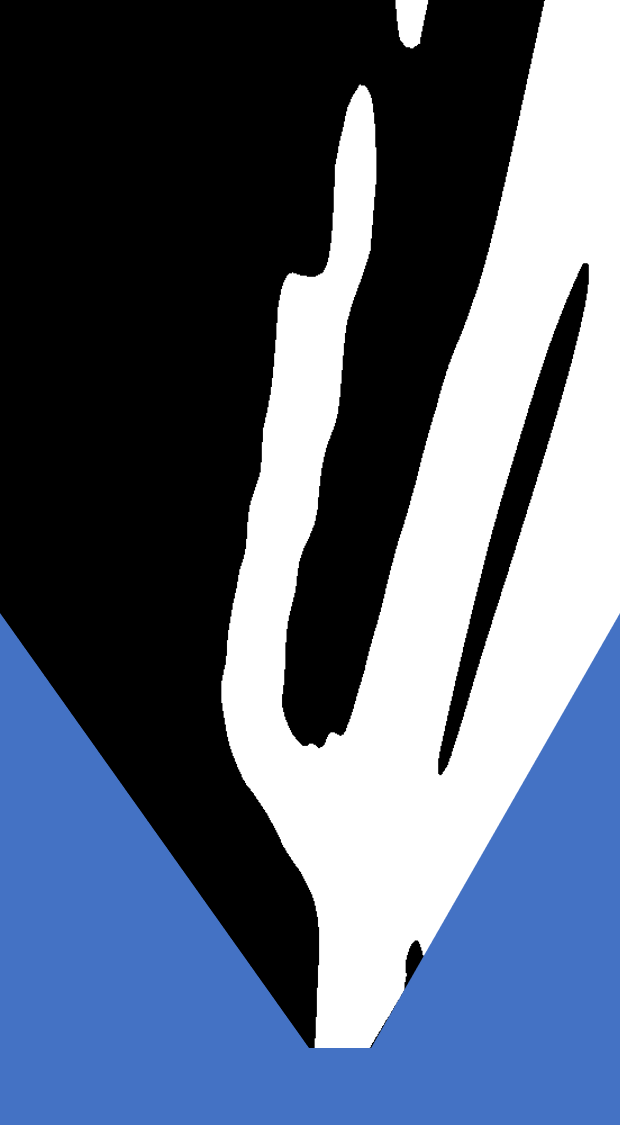}}&
{\includegraphics[width=0.12\textwidth]{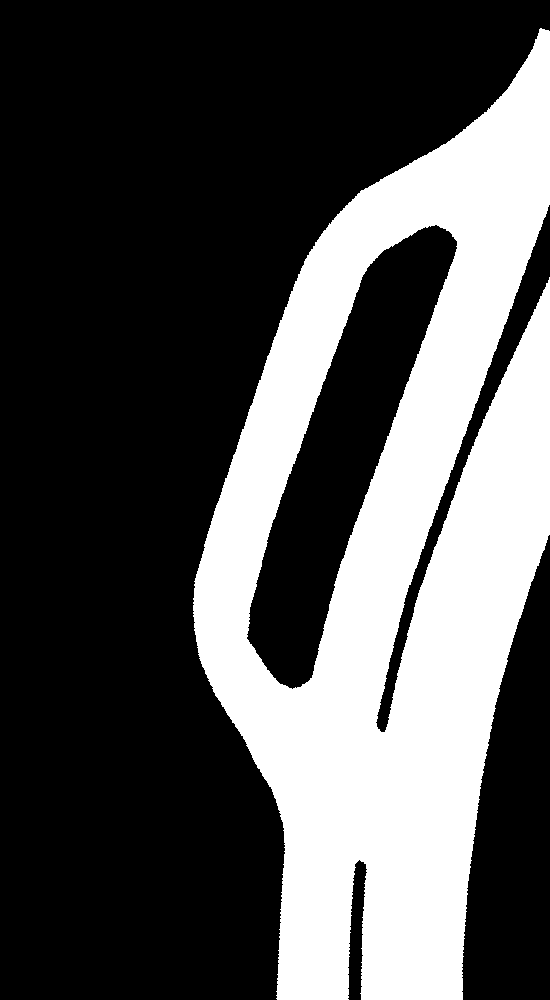}} &
{\includegraphics[width=0.12\textwidth]{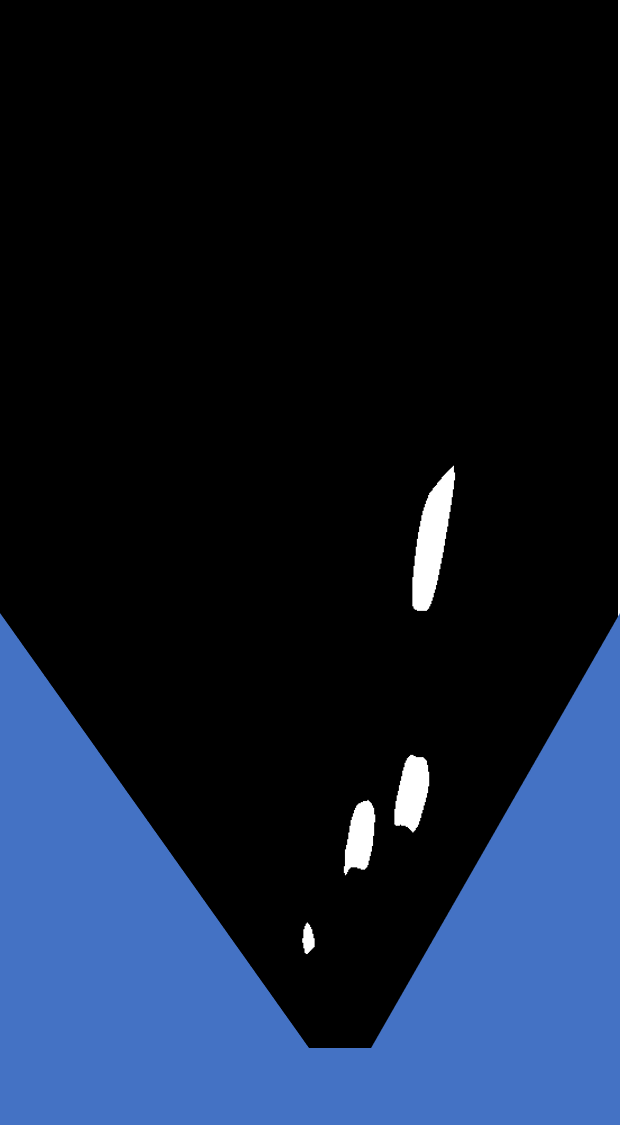}} &
{\includegraphics[width=0.12\textwidth]{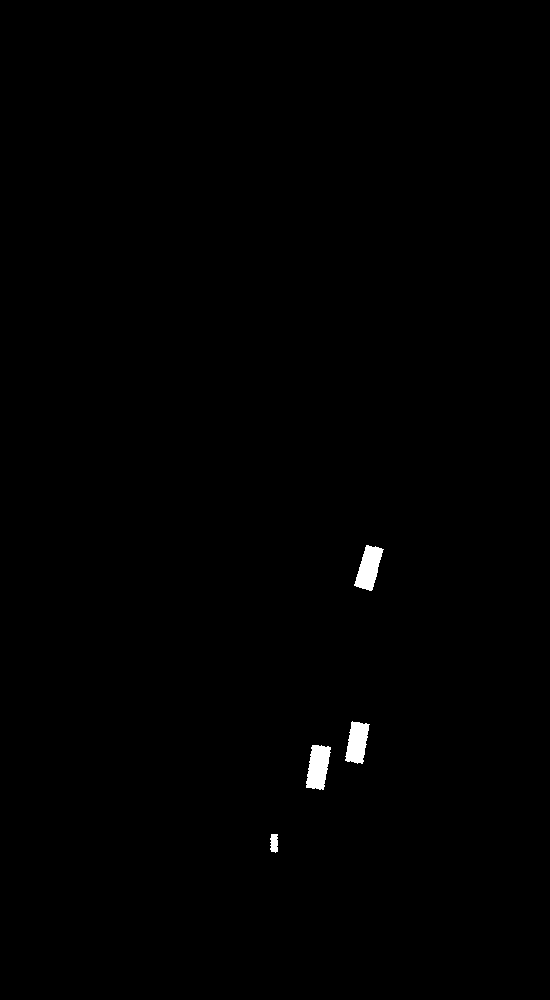}} \\
{\includegraphics[width=0.38\textwidth]{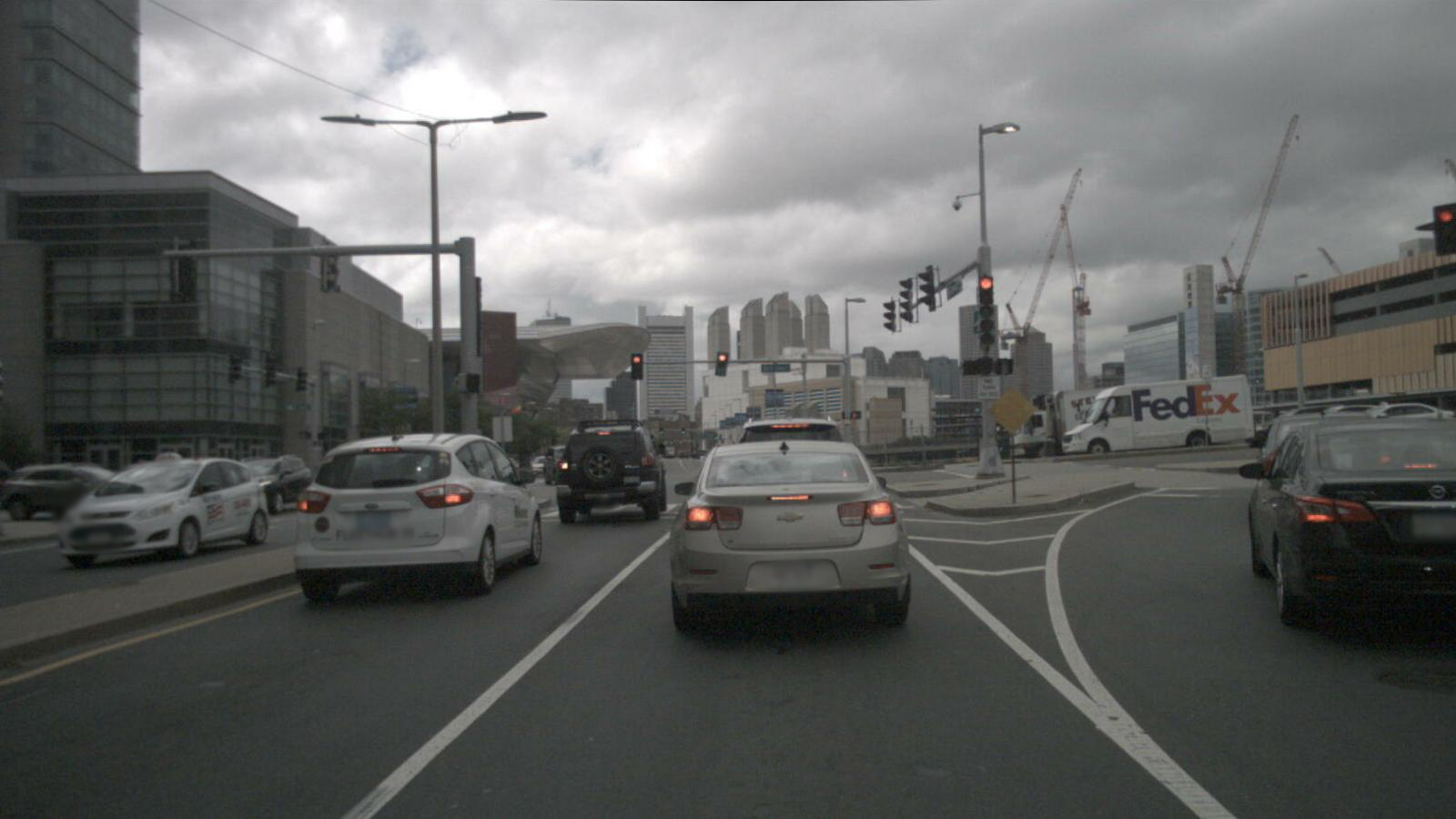}} &
{\includegraphics[width=0.12\textwidth]{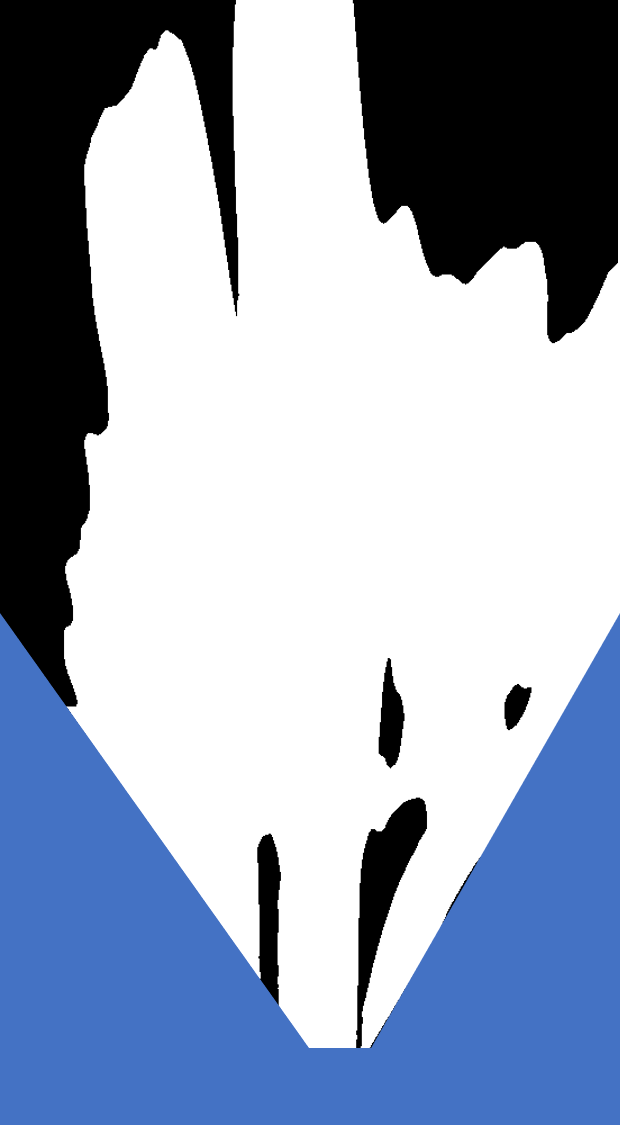}} &
{\includegraphics[width=0.12\textwidth]{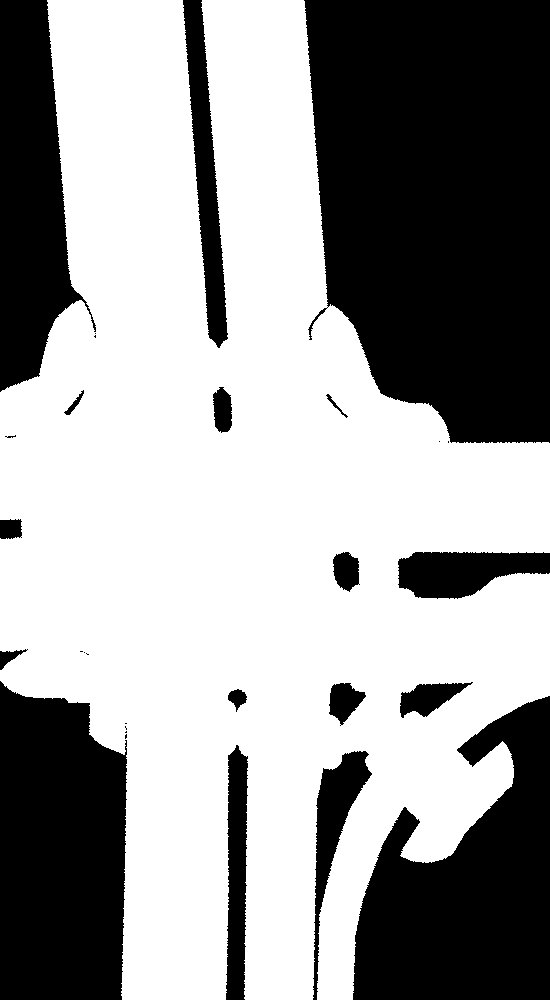}}&
{\includegraphics[width=0.12\textwidth]{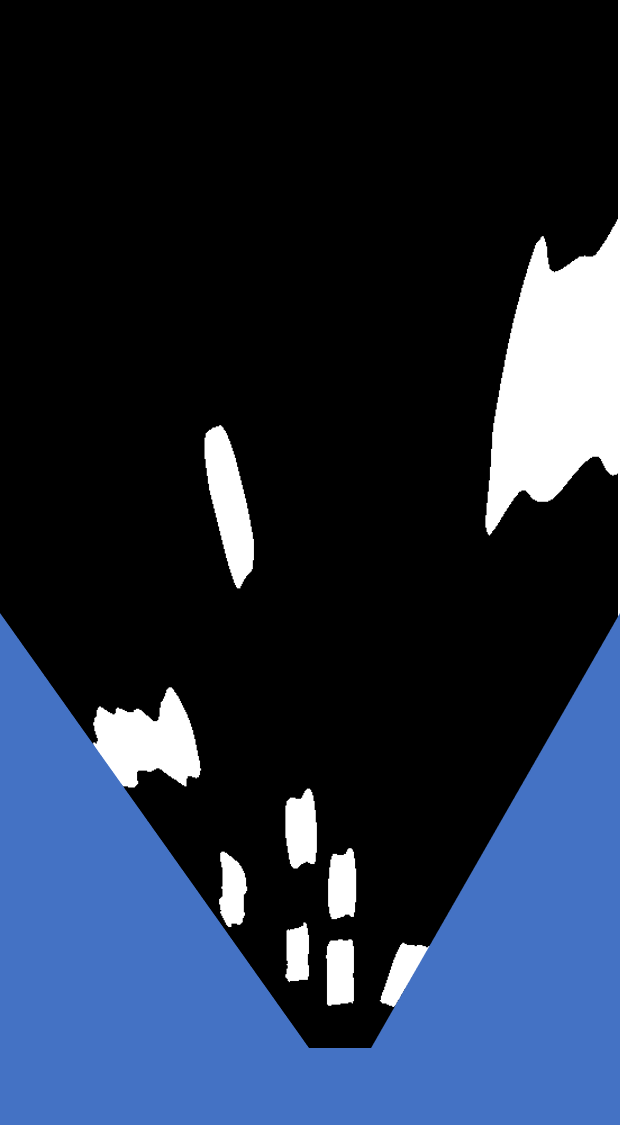}} &
{\includegraphics[width=0.12\textwidth]{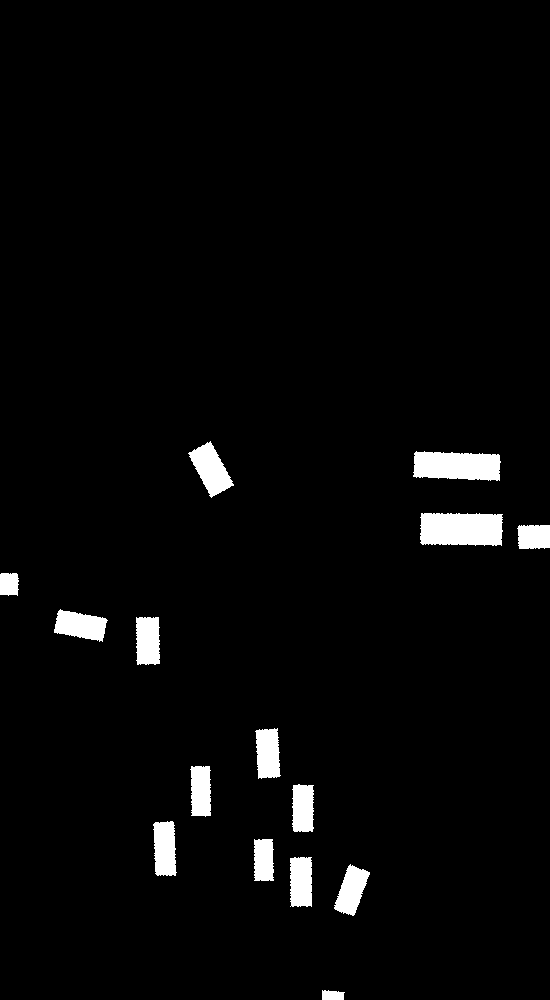}} \\
\multicolumn{1}{c}{RGB image} & Drivable area~(ours) & Drivable area (GT) & Vehicle (ours) & Vehicle (GT) \\[-2.75ex]
\end{tabular}
\end{center}
\caption{Bird-eye-view qualitative results for the first stage of the network. The blue part of the predicted masks corresponds to the limits of the camera's field of view. GT stands for Ground Truth. Better viewed in color.} \label{bvr}
\end{figure*}

Results presented in Table \ref{iou} show that our network outperforms OFT net \cite{roddick2018orthographic} by 5.4\% on the vehicle OGM and 2.5\% on the drivable area OGM. The gain in performance comes from the better accuracy in close range with an increase of more than 50\% on the drivable OGM and more than 25\% on the vehicle OGM. However, OFT has a better performance in long range: since it predicts directly in BEV, it does not suffer from the increase in error due to the imprecision in warping. 
Also, compared to OFT, our training criterion gives a higher relative weight to closer objects. Indeed, since our network learns in camera view, where closer objects appear bigger, it has an incentive to being more accurate in close range.
The same conclusions can be drawn from the comparison with VED \cite{Lu2018MonocularSO}. The BEV LiDAR baseline has a much better performance on the full OGM and in long range due to the 3D nature of the sensor but still falls behind our approach in close range.
A better performance in close range seems to be appropriate when planning a short-term trajectory. 
\vspace{-1.125ex}

\paragraph{Qualitative results}
Qualitative results for the first stage of the network are shown in Fig.~\ref{bvr}. As our network operates only on planar surfaces, warping the vehicle masks from camera view to BEV does not cause the ``stretching" that is observed in Fig.~\ref{is} when warping regular semantic masks without the knowledge of depth. Hence, our approach can distinguish two vehicles in front of each other by simply applying the connected components algorithm to the binary masks.
Again, we observe that results in BEV are better in the lower part of the OGMs, which corresponds to the first meters in front of the camera.
\vspace{-1.125ex}

\subsection{Trajectory planning}
\begin{table*}
\centering
\caption{Average displacement errors and $L1$ norm (in meters) for lateral and longitudinal displacements. NP (No Past) refers to models that do not take the past trajectory as input; CV (Camera view) refers to a model where  the predicted OGMs are not warped in BEV. Best results are shown in bold.}
\begin{tabular}{@{}l@{\hspace*{1.5ex}}cc*9{c}@{}}
\toprule
& \multicolumn{3}{c}{$ADE$}&\hspace*{1.5ex}&\multicolumn{3}{c}{$L1$ longitudinal}&\hspace*{1.5ex}&\multicolumn{3}{c}{$L1$ lateral} \\ 
\cline{2-4} \cline{6-8} \cline{10-12} \\[-2.5ex]
 & 0.5s & 1.5s  & 2.5s && 0.5s & 1.5s  & 2.5s && 0.5s & 1.5s  & 2.5s\\[-.5ex] \midrule
LSTM E-D  & 0.63 & 0.91 & 1.32 && 0.38 & 0.33 & 0.31 && 0.37 & 0.70 &  1.15  \\ 
End-to-end  & 0.54 & 0.84 & 1.21 && 0.32 & 0.34 & 0.28 && 0.34 & 0.61 &  1.04  \\ 
Mid-to-end  & 0.51 & 0.81 & 1.17 && \textbf{0.27}  & \textbf{0.26} & \textbf{0.19} && 0.32  & 0.63 & 1.04  \\ 
Holistic end-to-end CV  & 0.65 & 0.95 &  1.26  && 0.45 & 0.45 & 0.33 && 0.31 & 0.64 & 1.05\\
Holistic end-to-end (ours)  & \textbf{0.48}  & \textbf{0.78} & \textbf{1.16}  &&  \textbf{0.27} & \textbf{0.26} & 0.20 && \textbf{0.29} & \textbf{0.60}  & \textbf{1.02} \\ 
\midrule
End-to-end NP & 0.78 & 1.04 &1.40  && 0.64  & 0.52  & 0.42  && \textbf{0.30} & 0.68  & 1.14 \\
Mid-to-end NP & \textbf{0.65} & \textbf{0.94} & \textbf{1.25} &&  \textbf{0.44} & \textbf{0.44} & \textbf{0.32} && 0.31 & 0.62  & \textbf{1.03} \\ 
Holistic end-to-end NP  & 0.66  & \textbf{0.94} &  1.26 && 0.47  & 0.46 &  0.35&& \textbf{0.30} & \textbf{0.61} & \textbf{1.03} \\
\bottomrule
\end{tabular}
\label{ade}
\end{table*}
\paragraph{Evaluation baselines}
The performance of the holistic two-stage end-to-end model with intermediate OGM outputs is compared against the following baseline models:
\begin{itemize}
    \item LSTM E-D: Same architecture than the LSTM encoder-decoder described in Fig.~\ref{lstm}, except that it takes only the past trajectory as input.
    \item End-to-end: This model takes a sequence of camera images as input, encodes it with a ResNet-101, flattens the feature maps and feeds the obtained sequence of feature vectors to the same encoder-decoder LSTM described in figure \ref{lstm}. This model is comparable to direct perception approaches like \cite{Pomerleau1988ALVINNAA,DBLP:journals/corr/BojarskiTDFFGJM16,Codevilla2017EndtoEndDV} with the difference that it takes a sequence of images as input and leverages a LSTM model to process this temporal information. 
    \item Mid-to-end: Similar to the end-to-end model, except that it takes as input a sequence of concatenated {\em ground-truth} drivable area and vehicle semantic masks. This model is comparable to mid-to-mid approaches like \cite{DBLP:conf/rss/BansalKO19} or the privileged learner of \cite{chen2019lbc}.
\end{itemize}
\paragraph{Training setup}
All networks are trained using SGD with a batch size of 10 for 200 epochs. All results were obtained with a momentum of 0.9 and a learning rate of $10^{-3}$.
\paragraph{Quantitative results}
The results for motion planning are presented in Table~\ref{ade}. 
The $L1$ norm of lateral and longitudinal displacements are presented in addition to the Average Displacement error (ADE) metric.
The two-stage holistic network shows improvements on the three metrics over its regular end-to-end counterpart, mostly due to an improved accuracy in the longitudinal direction.
Overall, the regular end-to-end network has the worst performance among the image-based tested networks, which confirms the intuition that the intermediate BEV representation is an asset for motion forecasting.  
The mid-to-end network is accurate in the longitudinal direction but falls behind the holistic network in the lateral direction. 
A possible explanation for this surprising fact is that, our network learning the OGMs from camera images in an end-to-end fashion, it may benefit from pieces of information contained in the natural images which help disambiguate some driving scenarios. In other frameworks, accurate smooth estimates of discrete quantities have already been shown to be more efficiently processed than the ground truths themselves \cite{Hinton2015DistillingTK}. 
Our continuous masks, which are quite accurate in close range and camera view, may thus convey more information for fitting the second stage of the network in critical areas. The importance of having access to a scene context is highlighted by the inferior results obtained by the LSTM E-D model.
\vspace{-1.125ex}
\paragraph{Ablation study}
We show the importance of some components of the approach introduced here with the degraded results that are presented in Table~\ref{ade}.
Removing the past trajectory input is an important ablation to investigate the usefulness of the BEV when there is no prior about the past motion. An important gap in performance is observed between the holistic network that leverages BEV information to forecast motion and the end-to-end network with up to 17cm difference in short term longitudinal forecast, 7cm  in  long-term longitudinal forecast and 11 cm in long-term lateral forecast.
The importance of the BEV is also highlighted by removing the perspective warping layer from the holistic network architecture and feeding the Camera-View (CV) features as input to the second stage. The BEV version of the holistic network achieves a much better longitudinal performance than its CV counterpart with an average gain of 18cm at a horizon of 0.5s and 13cm at a horizon of 2.5s. However it is surprising to observe that the regular end-to-end network achieves a better performance than the holistic CV network. A possible explanation is that even if the RGB camera images are not optimal for motion forecasting, they still convey a better depth and perspective information than the camera-view semantic masks. 

\section{Conclusion}

In this paper, a novel method to output bird-eye-view occupancy grid maps from a monocular camera is introduced, by leveraging 3D bounding boxes and a HD-map raster.  
This method is used to build a new monocular camera-based holistic end-to-end motion planning network.
It takes as input a sequence of camera images and leverages the aforementioned bird-eye-view occupancy grid maps as an intermediate output.
Compared to its regular end-to-end counterpart, our method is better at imitation driving and provides interpretable intermediate results. 
The benefit of embedding computer vision tools for the transformation from the camera view to the bird-eye-view intermediate representation is also highlighted in our ablation study. 
Though end-to-end driving may not be reliable enough to be used as a primary solution in autonomous vehicles, having such a cheap camera based processing could be useful as an extra component for redundancy into a more complex system.

{\small
\bibliographystyle{ieee_fullname}
\bibliography{egbib}
}
\end{document}